# ProtDAT: A Unified Framework for Protein Sequence Design from Any Protein Text Description

Xiao-Yu Guo, Yi-Fan Li, Yuan Liu, Xiaoyong Pan, Hong-Bin Shen[*]

Institute of Image Processing and Pattern Recognition, Shanghai Jiao Tong University, and Key Laboratory of System Control and Information Processing, Ministry of Education of China, Shanghai, 200240, China

* Address correspondence to H.B. Shen at hbshen@sjtu.edu.cn
Tel: +86-21-34205320
Fax: +86-21-34204022

**Abstract：**

Protein design has become a critical method in advancing significant potential for various applications such as drug development and enzyme engineering. However, protein design methods utilizing large language models with solely pretraining and fine-tuning struggle to capture relationships in multi-modal protein data. To address this, we propose ProtDAT, a de novo fine-grained framework capable of designing proteins from any descriptive protein text input. ProtDAT builds upon the inherent characteristics of protein data to unify sequences and text as a cohesive whole rather than separate entities. It leverages an innovative multi-modal cross-attention, integrating protein sequences and textual information for a foundational level and seamless integration. Experimental results demonstrate that ProtDAT achieves the state-of-the-art performance in protein sequence generation, excelling in rationality, functionality, structural similarity, and validity. On 20,000 text-sequence pairs from Swiss-Prot, it improves pLDDT by 6%, TM-score by 0.26, and reduces RMSD by 1.2 Å, highlighting its potential to advance protein design.

**Introduction：**

Extensive research in the protein field has led to the accumulation of a large amount of multimodal data, supporting advancements in protein design. In recent years, models for protein structure prediction based on protein sequences have been continuously emerging, such as ESMFold [1], RoseTTAFold [2], and AlphaFold [3,4]. They bridge the gap between protein sequence and three-dimensional protein structure information, enabling the rapid acquisition of one modality (e.g., protein structure) from the other (e.g., protein sequence) [5]. The unimodal learning tends to constrain the performance of language models [6] since unimodal information can only partially capture the complexity of protein-related data, leading to an incomplete representation.

Protein design involves predicting or generating the sequences [7], structures [8], and functions [9] of proteins to address specific scientific and engineering challenges. Thus, the integration of multimodal data represents a significant trend in the development of protein design models. While the connection between protein sequence and structures are well established, the relationship between protein textual information (e.g., protein function) and protein sequences has become a focal point. Consequently, a variety of **P**rotein **L**anguage **M**odels (PLMs) are emerging, offering new perspectives and methods that enrich the protein design process.

The existing protein design methods based on PLMs can be broadly classified into 6 categories.

(1) **Only sequence modality.** ProGen2 [10] and ProtGPT2 [11] utilize millions of protein sequences for model pretraining. By leveraging the Transformer decoder [12] architecture, these models learn autoregressive sequence generation patterns, which can partially guide protein sequence generation.

(2) **Word prompts.** ProGen [13] and NetGo3.0 [14] perform large-scale pretraining by incorporating protein sequences along with text information. In ProGen, the word prompts are tokenized to regulate the generation of protein sequences.

(3) **Search models.** ProtDT [15] and ProtST [16] enhance the pretraining of protein sequences and text-based information through improvements in contrastive learning [17] and attention mechanisms. These models construct a multimodal interaction framework between protein description texts and protein sequences. While contrastive model is not capable of directly generating protein sequences, it can facilitate search functionalities. However, they are inherently limited and unable to handle entirely novel protein sequences [18].

(4) **Knowledge graphs.** OntoProtein [19] and PANNZER [20] integrate different modalities of data through knowledge graphs, designing networks to guide multimodal protein design.

(5) **Fine-tuning models.** Models such as ProLLaMA [21] and HelixProtX [22] incorporate multimodal protein information, including protein sequences, protein structures and descriptive annotations, into pretrained models. These models achieve modality interaction by designing abstractors between different modalities, enabling the precise generation of proteins.

(6) **Various types of data.** BioTranslator [23] and BioT5 [24] integrate manually annotated text with biological data, including drug and pathway information. This multimodal integration aids in protein design and supports various protein-related tasks.

However, there are several restrictions of existing PLMs. The emerging protein generation methods either focus solely on information from a single modality, such as protein sequences, or rely only on simple word prompts, lacking the ability to integrate other modalities to guide the protein generation process. Multimodal pretraining models for proteins present significant challenges, as they require high compatibility across various modalities, are susceptible to overfitting on training data, and encounter difficulties with modality alignment and expansion.

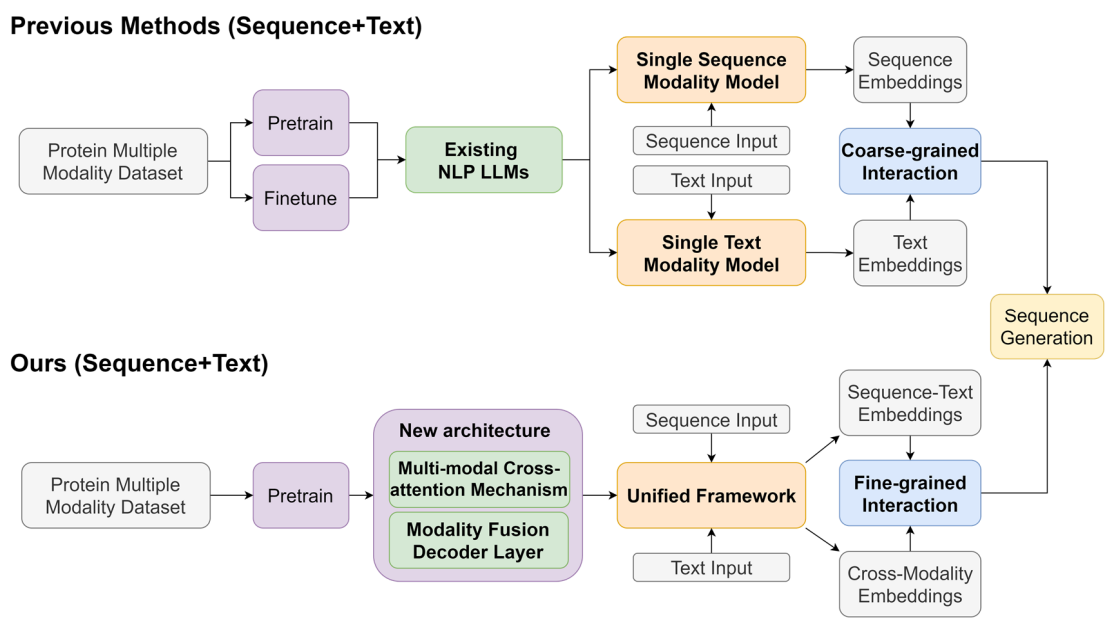

**Figure 1.** Previous methods generally pretrain or fine-tune single or multimodal data directly on **L**arge **L**anguage **M**odels (LLMs) without modifying internal structures like attention mechanisms, and they lack unified multimodal integration. In contrast, our method ProtDAT utilizes a **M**ulti-modal **C**ross-attention **M**echanism (MCM) to integrate multimodal data within a single framework, with foundational structural adjustments for each modality, optimizing the framework for text-guided protein generation.

Recently, generative models have been widely applied in protein design, achieving impressive results [25]. Previous work [26] has demonstrated that generating protein sequences from texts is effective and has been validated to some extent. In **Fig 1**, the training paradigm of previous generative models captures the correlation between protein sequences and their descriptive texts, allows for the expansion of knowledge, thereby enhancing the model's compatibility. However, these methods are only built on original LLMs by pretraining or fine-tuning for different modalities, enabling only coarse-grained interactions among outputs from different modality-specific models. Inspired by the large language models, such as the GPT [27] and LLaMA [28] series models, we developed ProtDAT (**Prot**ein **D**esign from **A**ny **T**ext Descriptions), a protein design framework that generates protein sequences based on any specified training dataset of textual descriptions. Our method designs Multi-modal Cross-attention Mechanism (MCM) for fine-grained interactions of different modality information at the modality fusion decoder layer. By altering the framework structure, it directly integrates embeddings internally, improving the efficiency and accuracy of text-guided protein generation.

To the best of our knowledge, ProtDAT is the first de novo fine-grained trained framework that addresses the significant issue of inadequate guidance from protein description texts for protein sequence generation in previous PLMs. Additionally, ProtDAT removes the limitations of prompt-based input, enabling training on any standardized protein dataset composed of protein sequences and descriptions. It pushes forward for the boundary of the long-standing challenge in biological data by building effectively linking to vast multimodal data, enabling organized natural language to be converted into new protein molecules with reasonable and novel sequence and structure. We first construct protein text-to-sequence datasets in a specific format and employ pretraining to develop a protein design model. ProtDAT surpasses the demand for fine-tuning across multiple pretrained models, introduces improvements to the cross-attention mechanism within Transformer models based on an autoregressive generation framework. These enhancements broaden the model's understanding of protein data and overcome the restrictions imposed by multimodal pretrained models in previous protein design frameworks.

ProtDAT is pretrained on the constructed protein text-to-sequence dataset (ProtDAT-Dataset) without explicit protein structure information. It employs a Transformer decoder-only architecture to facilitate interactions between the two modalities. A novel cross-attention mechanism is proposed in ProtDAT, i.e. MCM, which is specifically designed for handling comprehensive information, providing an intuitive, human-logical pathway to protein design. Unlike other PLMs that handle interactions between individual modal models, ProtDAT integrates the interaction of the two modalities within a single model, establishing a strong connection between them from the ground up. During this unsupervised training process, ProtDAT acquires the relationship between protein sequences and descriptions, analogous to how natural language models learn semantic and syntactic rules. This framework combines information from both modalities, enabling multimodal knowledge to interact throughout the entire training process, thereby allowing the model to generate protein sequences that satisfy specific goals.

In summary, the contributions of our research are as follows:

(1) We introduce ProtDAT, the first de novo fine-grained trained framework capable of being trained on any specialized protein text-sequence dataset. By addressing the barrier of inadequate guidance from protein descriptive texts in previous methods, ProtDAT enables the generation of entirely new proteins with broad applicability and flexibility.

(2) A novel cross-attention mechanism, MCM, is designed for the interaction between protein sequences and textual modalities, pushing the boundaries of protein design. Unlike previous approaches treating sequence and text as separate entities, MCM integrates both modalities at a foundational level, and builds effective linking to the transformation for multimodal data. It enables a unified approach that significantly enhances the model's ability to generate proteins that align more closely with the given textual descriptions.

(3) Experiments demonstrate that in the protein generation process, textual information plays a dominant role with MCM module. ProtDAT achieves the state-of-the-art results across various evaluation metrics for both sequence and structure on the Swiss-Prot [29] dataset.

## Results:

**ProtDAT Training Framework and Data Preparation**

The ProtDAT framework in **Fig 2a** consists of three primary input components: protein description text, protein sequence, and cross-modality tensor. These multimodal inputs are first processed by a preprocessing module that vectorizes the information. The vectorized embeddings are then passed through decoder layers with the MCM module, which enables the model to generate modality-specific output vectors. Finally, these vectors are utilized in downstream protein generation tasks, facilitating the synthesis of novel protein sequences based on the integrated textual and sequence information.

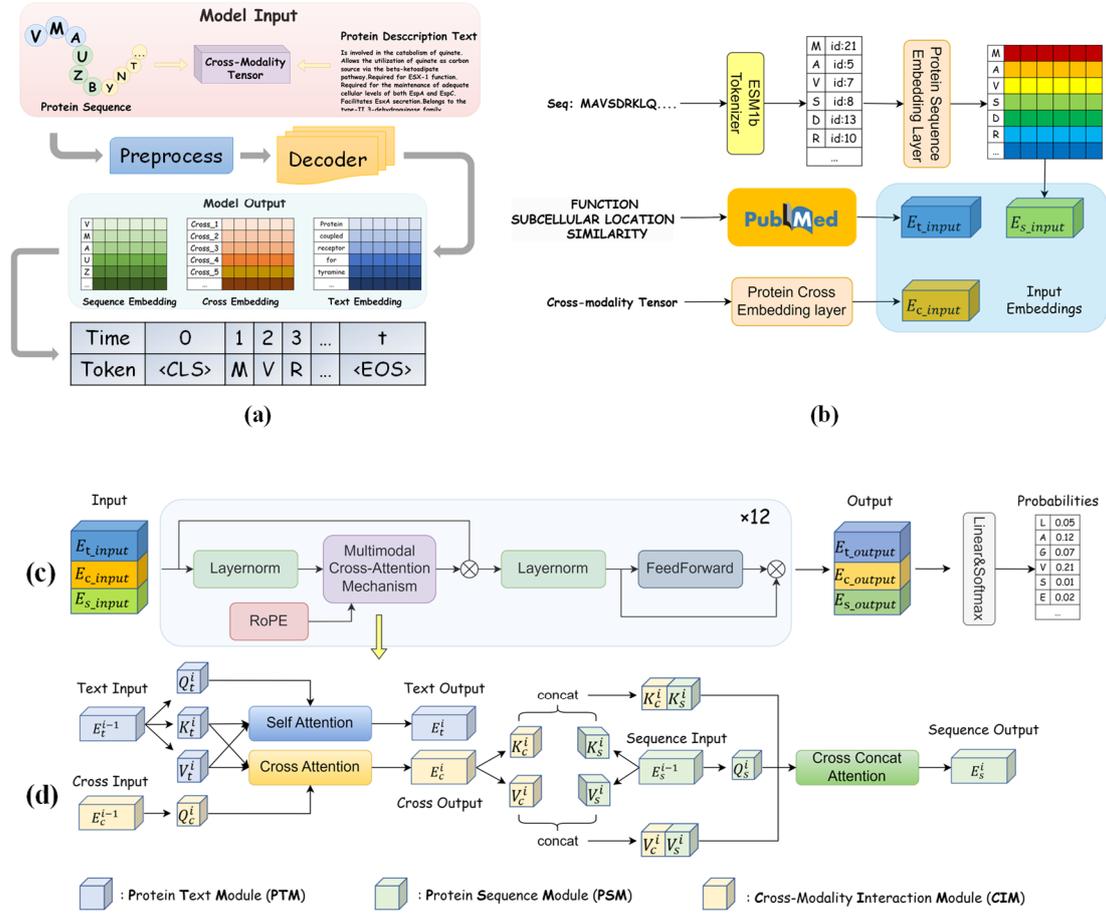

**Figure 2. The diagram of ProtDAT framework. (a)** The components of the ProtDAT framework, including model input, preprocess model, decoder layers, model output and sequence generation module. **(b)** Preprocessing of the ProtDAT-Dataset training dataset, which includes three components: protein sequences, protein descriptive texts, and modality cross vectors. **(c)** Pretraining process of ProtDAT, where the model receives the output from **Fig 2a** through a decoder-only architecture, generates entirely new protein sequences. **(d)** Flowchart of the MCM module, which applies self-attention and cross-attention mechanisms to the PTM, CIM, and PSM, respectively, ultimately producing the outputs of the current module.

The ProtDAT framework in **Fig 2c** includes 12 decoder layers, each containing layer normalization, **Ro**tary **P**osition **E**mbedding (RoPE) [30], MCM, and a feedforward network. Embeddings of the multimodal inputs in **Fig 2b** are processed through decoder layers to generate corresponding output vectors. They are then transformed via a linear mapping and a Softmax function to produce probability vectors of protein sequences.

As previous autoregressive methods, ProtDAT is also trained with a cross-entropy loss function.

ProtDAT leverages MCM, specially designed to integrate protein sequences with their descriptive texts, as shown in **Fig 2d**. Unlike conventional cross-attention approaches, MCM employs a **C**ross-**M**odality **I**nteraction **M**odule (CIM) to integrate information from the **P**rotein **S**equence **M**odule (PSM) and the **P**rotein **T**ext **M**odule (PTM), enhancing learning between modalities. Built upon **M**ulti-**H**ead **A**ttention (MHA), MCM incorporates self-attention, cross-attention, **C**ross-**C**oncat-**A**ttention (CCA) mechanisms and corresponding custom masking strategies to ensure controlled interaction and prevent error accumulation in generation tasks. This approach overcomes limitations in previous models that treated text annotations and sequences as separate modules, integrating diverse modality data seamlessly within a unified framework.

We constructed the ProtDAT-Dataset shown in **Table A1** of **Appendix A** by obtaining a non-redundant set of 469,395 protein entries from the well-annotated Swiss-Prot [29] database, with each entry consisting of a protein description paired with its corresponding protein sequence. The protein description texts include information on protein function, subcellular localization, and protein family, which are embedded utilizing PubMedBERT [31], while the protein sequences are tokenized with ESM1b [32]. ProtDAT-Dataset is partitioned into 402,395 entries for the training set, 47,000 entries for the validation set, and 20,000 entries for the test set. Additionally, ProtDAT does not enforce rigid formatting requirements on the dataset, allowing for flexible construction based on task-specific description content.

**ProtDAT Generates Protein Sequences from Multi-modal Data**

To evaluate the effectiveness of protein sequences generated by ProtDAT, we utilize the test set of ProtDAT-Dataset, containing 20,000 text-sequence pairs, as a reference dataset. ProtDAT employs a generation strategy that combines a temperature coefficient, Top-p decoding, and a repetition penalty to separately enhance diversity, reduce errors, and prevent repetitive sequences. This probabilistic sampling approach [33] draws multiple samples at the start of generation, further optimizing the diversity and orderliness of the generated sequences. Therefore, we further develop three prompt methods for Progen2 [10], ProtGPT2 [11], ProLLaMA [21] and ProtDAT, which perform the following protein generation tasks.

**Prompt method 1 (PM1).** Only the protein description texts corresponding to the reference protein sequence are utilized as prompts during the generation process.

**Prompt method 2 (PM2).** Based on PM1, the first 10 amino acids of corresponding sequence are added, with both the sequence and text modalities jointly guiding the generation process.

**Prompt method 3 (PM3).** Only the first 10 amino acids of the reference protein sequence are used as the prompts in the generation process.

**Prompt method 4 (PM4).** Building on PM3, the subcellular localization information from the protein annotations corresponding to the reference sequences is incorporated.

**Table 1. Details of Generated-Dataset.** The protein sequences generated by Progen2, ProtGPT2, ProLLaMA, and ProtDAT, utilizing their respective prompt methods to constitute the Generated-Dataset with a total of 90,805 sequences.

| Method | Coherent Text[a] | Prompt Method[b] | Prompt Modality[c] | Seq Count[d] |
|---|---|---|---|---|
| Progen2 | / | PM3 | Sequence | 20000 |
| ProtGPT2 | / | PM3 | Sequence | 20000 |
| ProLLaMA | √ | PM4 | Text+Sequence | 10805 |
| ProtDAT | √ | PM1 | Text | 20000 |
| ProtDAT | √ | PM2 | Text+Sequence | 20000 |

[a] The ability to process coherent natural language descriptions of proteins.
[b] The prompt method of each model or framework.
[c] The modality (modalities) of corresponding prompt method.
[d] The number of generation protein sequences of each model or framework.

Each of Progen2 and ProtGPT2 generates 20,000 protein sequences by applying PM3, as they lack the ability to process coherent textual input. While ProLLaMA adopts a dual-modality approach using PM4, it only incorporates protein superfamily information. Therefore, 10,805 protein pairs are extracted from the ProtDAT-Dataset that match the protein superfamilies dataset of ProLLaMA. We used these superfamily-sequence pairs as prompts to guide the protein generation process. Our framework, ProtDAT, employs PM1 and PM2 on the same test set, generating a total of 40,000 protein sequences—20,000 sequences for each method. We refer to the entities of all the generated protein sequences as Generated-Dataset, illustrated in **Table 1**.

For the protein sequences generated by each method, we employ ESMFold to obtain their structures, which are compared them with the protein structures in reference dataset. The following metrics are leveraged to evaluate the generated proteins.

(1) **Global sequence identity**, which intuitively evaluates the similarity between the generated and original sequences through global protein sequence alignment.

(2) **pLDDT** [3], which is used to represent the confidence level of the structural prediction for each amino acid, mainly reflecting the reliability of individual residues.

(3) **TM-score** [3], which measures the similarity between two protein structures and is more sensitive to overall shape changes, making it suitable for comparing proteins of different lengths.

(4) **RMSD** [34], an indicator that compares the atomic coordinate differences between two structures, is used to quantify the discrepancy between the predicted structure and the reference structure.

Except for RMSD, where lower values indicate smaller differences, all other metrics are better when higher. These evaluation metrics for sequences in Generated-Dataset are shown in **Table 2** and **Fig 3**. The pLDDT values are directly obtained from the structural **P**rotein **D**ata **B**ank (PDB) [35] files, while TM-score and RMSD values are acquired through TM-align [36].

**Table 2. Evaluation metric results.** Evaluation results of the Generated-Dataset which is built by sequences generated by Progen2, ProLLaMA, ProtGPT2, and ProtDAT under two different prompt methods across various metrics [a].

| Method | Seq-Identity | pLDDT↑ | TM-score↑ | RMSD↓ |
| --- | --- | --- | --- | --- |
| Progen2 | 0.207 | 58.64 ± 23.90 | 0.344±0.238 | 5.494±2.060 |
| ProLLaMA | 0.212 | 41.71 ± 15.43 | 0.228±0.080 | 5.536±0.927 |
| ProtGPT2 | 0.183 | 57.56 ± 23.11 | 0.240±0.188 | 4.601±2.066 |
| **ProtDAT(PM1)** | 0.304 | **64.41 ± 25.86** | 0.598±0.300 | 3.509±2.223 |
| **ProtDAT(PM2)** | **0.334** | 63.97 ± 25.82 | **0.607±0.300** | **3.477±2.234** |

[a] Protein structure files are obtained by ESMFold. Except for global sequence identity, all other metrics are presented as mean ± standard deviation, calculated based on the total number of proteins generated by the different models and frameworks in Table 1.

In **Fig 3a**, the maximum length of generated sequences for each method is set to 500 tokens. Given that the average length of sequences in the ProtDAT-Dataset test set is 387 tokens across 20,000 sequences, results indicate that while most methods can reach this maximum length, sequences generated by ProLLaMA are noticeably shorter and do not achieve the full 500 tokens. As the length of generated sequences varies in **Fig 3a**, the evaluation metrics of protein sequences and structures remain stable across various lengths, and there is no decline in confidence as sequence length of ProtDAT increases. Leveraging the complete and coherent protein description text as the sole prompt allows the framework to generate sequences that more closely resemble natural proteins. With the additional content of protein sequence prompts, it can further improve the quality of generated sequences and reduce error accumulation.

In **Table 2**, ProtDAT achieves the best performance across all metrics under both prompt methods, with average values of the global sequence identity at 0.334, pLDDT at 64.41, TM-score at 0.607, and RMSD at 3.477. In **Fig 3b**, pLDDT shows an average improvement of 6%, while in **Fig 3c and Fig 3d**, it is evident that the structural similarity of ProtDAT shows a significant improvement. The confidence in structural prediction shows a steady improvement, especially with the structural similarity metric TM-score, which increased by approximately 0.26 on average, while RMSD decreased by more than 1.2 Å. These metrics indicate that the protein sequences generated by ProtDAT align reasonably well with the detailed descriptions from the text, even when the global sequence identity is low.

**Fig 3e** reveals that, even when the global sequence identity is quite low between 20-40%, TM-score remains relatively high from 0.8 and 1.0. This indicates that ProtDAT has effectively learned the relationship between protein sequences and textual information, enabling it to generate novel proteins with low sequence but high structural similarity. **Figure 3f** shows that the generated proteins generally exhibit high values for both metrics, indicating not only high sequence similarity but also high confidence in the generated proteins. Most of the generated sequences have a pLDDT score between 80% and 100%, and a TM-score ranging from 0.8 to 1.0. More evaluation metrics can be found in **Appendix E**.

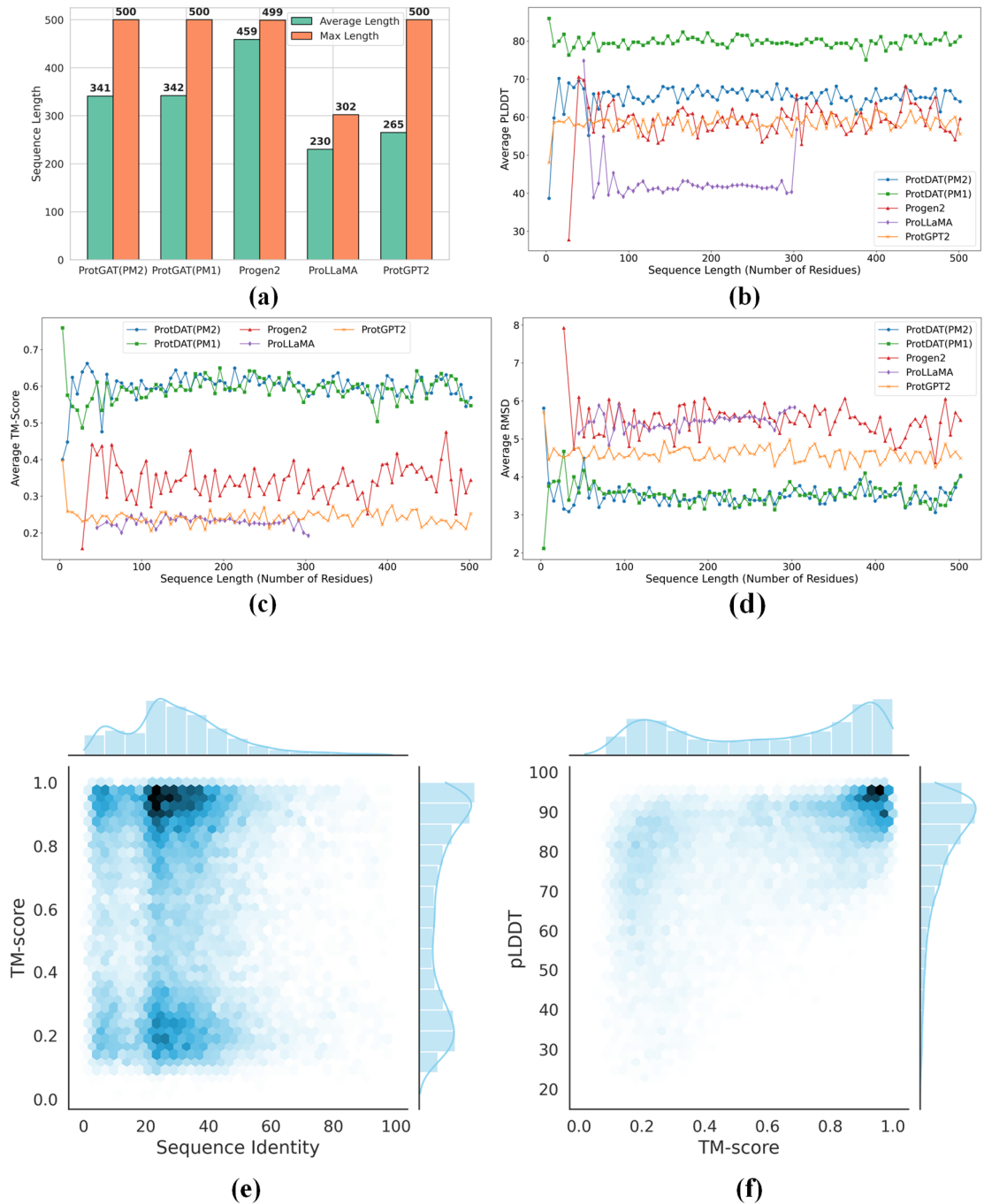

Figure 3. **Various evaluation metrics based on generated protein sequences of different methods**. **(a)** A bar chart showing the average and maximum lengths of generated sequences by different methods (with a maximum length limit of 500). **(b)**, **(c)**, and **(d)** individually display the mean plDDT, TM-score, and RMSD values of generated proteins by the above methods, with the sequence length ranging from 1 to 500. In **(e)** and **(f)**, we use 20,000 protein sequences generated with PM1 prompts from the ProtDAT-Dataset test set, while darker colors typically indicate higher data density, meaning more data points in that area. **(e)** The distribution of global sequence identity and TM-score. **(f)** The distribution of TM-score and pLDDT.

## MCM's Role in ProtDAT Sequence Generation

In **Fig 2**, MCM proficiently integrates PSM, PTM, and CIM, enabling ProtDAT to generate protein sequences guided by description texts. To further illustrate the role of different attention mechanisms in facilitating the sequence generation within MCM, we visualize the attention weights across different modules during the protein sequence generation process. we applied to the 20,000 generated sequences using PM1 prompts of ProtDAT as described in **Table 1**.

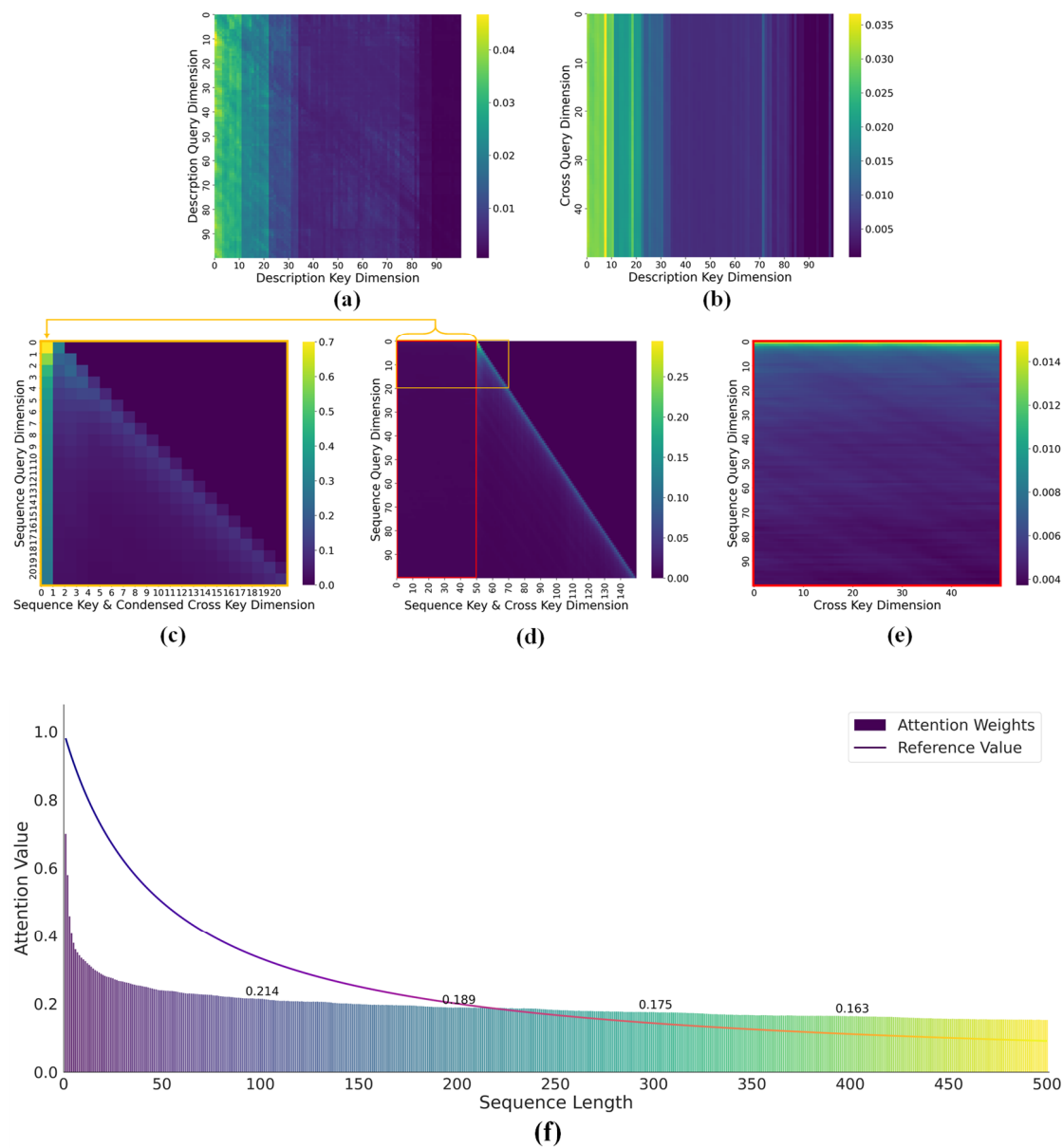

**Figure 4. Attention heat maps of MCM in ProtDAT.** The attention weight diagrams are presented using at most 100 tokens for visualization purposes, and the token length depends on the specific protein description text and sequence length. **(a)** The self-attention mechanism weight diagram for text embeddings of PTM. **(b)** The cross-attention weight diagram between cross embeddings of CIM and text embeddings of PTM. **(c)** The the magnified views of the yellow specific regions in (d), while the CIM attention weights are condensed in one dimension. **(d)** The CCA weight diagrams between sequence embeddings of PSM and cross embeddings of CIM. **(e)** The magnified views of the red specific regions in (d), which demonstrate the details of CIM attention weights in sequence modality fusion. **(f)** The proportion of the overall attention weight contributed by CIM in the protein sequence generation procedure, as shown in the CCA weights in **Fig 2d.** While the curve is the reference line of attention weights, showing a decreasing trend as the sequence is generated.

Assume $\mathcal{D}_t, \mathcal{D}_c, \mathcal{D}_s$ represent the lengths of the text, cross-modality tensor, and sequence, respectively, the shape of attention weights in **Fig 4a** and **Fig 4b** are $\mathcal{D}_t \times \mathcal{D}_t, \mathcal{D}_c \times \mathcal{D}_t$. Additionally, to highlight the impact of cross-modality tensor in the CCA mechanism, its attention weights in **Fig 4d** is merged. Thus, the attention weights originally have a dimension of $\mathcal{D}_s \times (\mathcal{D}_c + \mathcal{D}_s)$ to illustrate the influence of CIM and PSM in guiding protein sequence generation. Therefore, with $\mathcal{D}_c$ is condensed into one dimension, the shape of attention weights in **Fig 4c** is effectively represented as $\mathcal{D}_s \times (1 + \mathcal{D}_s)$. It provides a magnified view of the first 20 tokens from **Fig 4d**, allowing for a clearer representation of the weights of CIM in guiding sequence generation. Also, **Fig 4e** is a zoomed-in view of the first $\mathcal{D}_c$ columns of attention weights in **Fig 4d**. To facilitate visualization analysis, at most 100 dimensions of each modality can be displayed.

**Fig 4** visualizes the average attention weights of the three attention mechanism modules in MCM during the generation process. The self-attention mechanism in **Fig 4a** is similar to the training process of the BERT [37] based model, appropriately integrating single modality data. The attention weights of the protein description text tokens gradually decrease as the text length increase, indicating that textual tokens in the beginning positions have more substantial influence in guiding the generation method. In **Fig 4b**, the weight differences for each token in PTM are not significant, indicating that the cross-attention mechanism effectively integrates the textual information into the cross embedding (with $\mathcal{D}_c$ is 50), laying the foundation for subsequent fusion with protein sequence information.

In **Fig 4d**, the CCA module is performed under a causal mask shaped $\mathcal{D}_s \times \mathcal{D}_s$ only apply to the last $\mathcal{D}_s$ columns of CCA attention weights, where it can be observed that the selection of each amino acid is closely related to the weights of CIM and the preceding few amino acid tokens of PSM. As illustrated in **Fig 4c,** the compact attention weights of CIM play a significant role in protein sequence generation, holding a substantial weight and serving as a critical part of the procedure. Furthermore, the weights of few amino acid tokens just before the current generation time step are relatively high. The results indicate that error accumulation is a persistent phenomenon when using autoregressive models for designing proteins. The inclusion of CIM in generation procedure addresses this issue to some extent, stabilizing the whole sequence generation process.

**Fig 4e** is another detailed view of **Fig 4d** with shape as $\mathcal{D}_s \times \mathcal{D}_c$. The attention weights decrease to some extent as the dimension of $\mathcal{D}_s$ (i.e., the number of rows) increases, indicating that the guidance of the protein description text on sequence generation diminishes as the protein sequence lengthens, while still maintaining a certain level of influence.

Thus, **Fig 4f** provides a more intuitive display of average attention weights in **Fig 4e** in facilitating the generation process as sequence length (maximum to 500) grows. Assuming using $\mathcal{D}_c$ amino acid tokens to assist ProtDAT in generation, a function is derived as $\mathcal{D}_c * 1/(\mathcal{D}_c + m)$ to demonstrate the weights of the prompt tokens, where $m$ denotes the length of already generated sequence tokens in the procedure. This function is presented as a curve named 'Reference Value' in **Fig 4f**. It is observed that as the protein sequence length rises, the contribution of the protein sequence portion used as a prompt to subsequent sequence generation diminishes. However, in our method, when the sequence length reaches 100, the contribution of the protein descriptive text to the generation stabilizes at around 20%, regardless of the sequence length. This further demonstrates that the textual description ensures the accurate instruction of essential amino acid tokens during the early stages of generation, and continues to provide directional support as the protein sequence grows. This effectively addresses the issue of instability in generation as the protein sequence lengthens.

In conclusion, MCM plays a critical role in guiding the protein generation. As a bridge connecting protein description texts and protein sequences, CIM significantly contributes to both the integration of information from protein texts and the guidance of sequence generation. MCM tackles the problem of insufficient guidance from protein descriptive texts in existing methods, facilitating robust multimodal integration and offering a cohesive framework to handle both modalities. By leveraging MCM, ProtDAT greatly enhances the precision of generated protein sequences, ensuring better alignment to the provided textual descriptions.

**ProtDAT Designs Remote Homologs Protein Sequences**

To further investigate whether the protein sequences generated by ProtDAT possess the characteristics described in the text, we compared 40,000 sequences generated from ProtDAT with different prompt methods in Generated-Dataset with the test set of ProtDAT-Dataset. ESM1b has extensively learned the representation patterns of protein sequences, it produces protein representation vectors that encode diverse structural and functional properties. These vectors can be further analyzed and compared in the vector space, enabling a deeper understanding of the relationships among proteins through the dimensionality reduction technique *t*-SNE [38] and clustering algorithm K-means [39],. This approach facilitates not only protein classification but also the prediction of novel functional insights by exploring the vector space for hidden patterns.

In **Fig 5a**, it is evident that the distribution of the generated and reference protein sequences in the vector space exhibits a significant similarity, indicating that the generated sequences share similar underlying properties with the test set. Notably, the clustering of sequences is predominantly concentrated at the bottom of **Fig 5a**, suggesting a high density of points in this region. This pattern of distribution further emphasizes the model's ability to capture and replicate the spatial characteristics of sequences in the embedding space, reinforcing its effectiveness in generating biologically relevant protein representations.

To further assess the level of similarities between the generated sequences and their corresponding descriptive texts, we selected six protein sequences from different regions in the mapped space with UniProt [40] IDs. These generated protein sequences exhibit an average structural similarity of over 90%, while maintaining a sequence similarity less than 25%. Such characteristics align with the definition of remote homologs, which are proteins with low global sequence similarity but high structural resemblance. This result highlights the capability of the generation process to explore and replicate the principles of remote homolog evolution, producing novel sequences that deviate at the sequence level while significantly preserving structural conservation.

The protein generation and evaluation process of ProtDAT is outlined as follows, with **Fig 5b** illustrating one case. First, ProtDAT is informed of the generation task through the input of different prompt methods, followed by the selection of generation parameters to produce protein sequences. Subsequently, the generated sequences are subjected to a similarity analysis against the references to obtain global sequence similarity. Additionally, the natural and generated sequences can be input into protein structure models such as ESMFold to derive their structures. These structures are then compared using TM-align to assess TM-score, pLDDT, and RMSD, which measure structural quality and similarity.

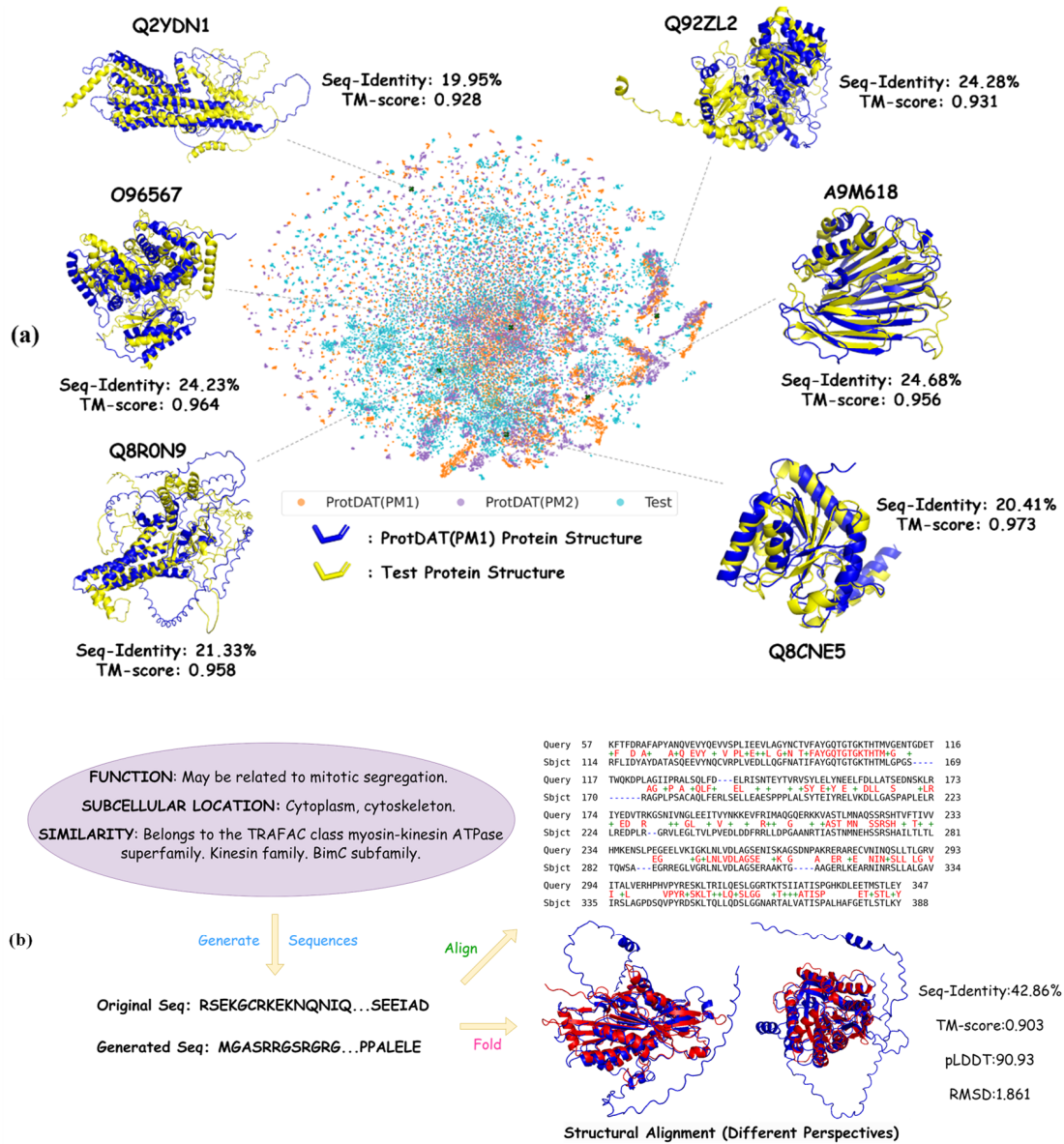

**Figure 5**. **Protein space visualization.** **(a)** The point cloud distribution of protein vectors. 'ProtDAT(PM1)', 'ProtDAT(PM2)', and 'Test' represent protein sequences generated using prompt method 1 and prompt method 2, as well as the reference protein sequences from ProtDAT-Dataset test set, respectively. The structures of the reference and generated sequences are depicted separately in yellow and blue, with the UniProt IDs and results of global sequence identity and Tm-score. **(b)** A case of ProtDAT design process with PM1, comprises four components: prompt input, protein sequence generation, protein sequence alignment, and protein evaluation.

**ProtDAT generate sequences with natural protein characteristics**

In the ProtDAT generation module, whether the generated protein sequences closely resemble the natural arrangement of amino acids is a crucial evaluation criterion. Therefore, designing novel sequences is of paramount importance.

To determine the values of the generation parameters, Optuna [41] hyperparameter optimization framework is deployed, randomly selecting 3,000 protein sequences of the ProtDAT-Dataset test set. The similarities between generated sequences and natural sequences are calculated by KL divergence [42], thereby determining the range of generation parameters, as shown in **Fig 6a**. Subsequently, the model generates sequences via protein description texts under different Top-p and temperature coefficient parameters, while Top-p filters out tokens with low probabilities and temperature coefficient controls the diversity in the sequence generation process. Since the global sequence identity [43] cannot directly determine whether protein sequences have similar functions, therefore TM-score is implemented to evaluate the generation results from the 3D structures' perspective. Experiments are conducted with the repetition penalty set to 1.2 according to our local tests, it prevents excessive repetition of tokens, ensuring sequence variability.

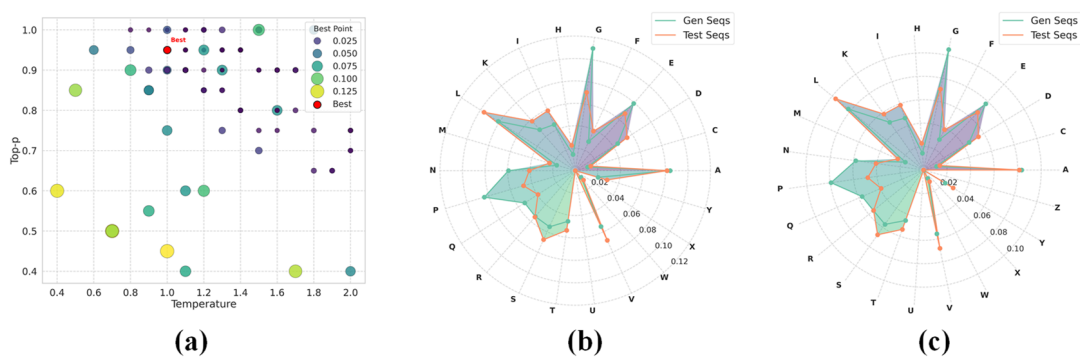

(a) (b) (c)

**Figure 6**. **Similarities between generated and natural protein sequences.** **(a)** The similarities between the protein sequences generated by ProtDAT under different generation parameters and natural protein sequences, calculated by KL divergence. **(b) (c)** The amino acid residue distribution of protein sequences generated separately in PM1 and PM2 under the conditions of Top-p=0.85 and T (temperature coefficient) =1.0, compared to the corresponding natural sequences.

In **Fig 6a**, the optimal settings were identified as Top-p=0.95 and T=1.0. Therefore, we selected Top-p values ranging from 0.55 to 1.0 with a step of 0.15 and temperature coefficients from 0.4 to 1.4 with a step of 0.2 for further generation experiments. The TM-Vec model [44] was employed to rapidly assess structural similarity from the generated protein sequences. In **Table 3**, the TM-score reaches the highest value when Top-p is 0.85 and the temperature coefficients are 1.0 and 1.2. To introduce randomness in sequence generation, the temperature coefficient in the subsequent model generation processes is set to 1.0.

**Table 3**. **TM-score of generated sequences with different parameters.** TM-score values are obtained with the TM-Vec model, representing the structural similarity between the sequences generated by ProtDAT in PM1 of ProtDAT-Dataset under different generation parameters.

| TM-score | T=0.4 | T=0.6 | T=0.8 | T=1.0 | T=1.2 | T=1.4 |
|---|---|---|---|---|---|---|
| **Top-p=0.55** | 0.5832 | 0.5911 | 0.6020 | 0.6104 | 0.6120 | 0.6074 |
| **Top-p=0.70** | 0.5987 | 0.6032 | 0.6204 | 0.6430 | 0.6408 | 0.6316 |
| **Top-p=0.85** | 0.6039 | 0.6114 | 0.6307 | **0.6502** | **0.6502** | 0.6399 |
| **Top-p=1.0** | 0.6045 | 0.6144 | 0.6297 | 0.6482 | 0.6500 | 0.6403 |

With generation parameters of Top-p=0.85 and T=1.0, and a repetition penalty of 1.2 applied consistently across all tasks, ProtDAT achieves randomness in protein sequence generation, reducing excessive consecutive repetitions of amino acids and minimizing low-probability tokens, thereby enhancing the generation accuracy. As shown in **Fig 6b** and **Fig 6c**, the generated protein sequences from ProtDAT exhibit a high degree of similarity with the amino acid sequence distribution of natural proteins, indicating that the model has effectively learned the specific patterns of text-guiding protein generate process.

## Conclusion and discussion

Proteins can be described through various modalities, including natural language text, sequences, and structures, making the integration of different modalities crucial for a comprehensive understanding. ProtDAT stands as the first de novo fine-grained trained framework that can be trained on any specialized protein text-sequence dataset. By overcoming the persistent challenge of insufficient guidance from protein descriptive texts in previous methods, ProtDAT enables the generation of entirely new proteins with wide-ranging applicability and flexibility. We introduce MCM, a novel cross-attention mechanism specifically designed to facilitate interactions between protein sequences and textual data. Unlike existing methods that handle sequence and text independently, MCM integrates both modalities from the ground up, establishing robust connections for multimodal data transformation. Experimental results demonstrate that the protein sequences generated by ProtDAT effectively incorporate text information, achieving promising performance in terms of rationality, functionality and structural similarity with an average 6% improvement in pLDDT, a 0.26 increase in TM-score, and a 1.2 Å reduction in RMSD. Relevant resources can be found at https://github.com/GXY0116/ProtDAT.

In the future, we plan to pretrain ProtDAT by expanding its linguistic capacity with a wider range of annotated protein datasets for more nuanced text-guided protein generation. Additionally, we aim to broaden the MCM by incorporating structural attention mechanisms, extending current protocol to cover more modality, creating a more powerful tool for the field of protein design. Furthermore, we intend to extend ProtDAT beyond protein sequences to other biological languages, such as RNA, drug design and single-cell data, by training the framework on datasets from different domains with advanced mechanisms.

## Methods

**Construction and Preprocessing of ProtDAT-Dataset.**

To train the ProtDAT, we obtained a non-redundant set of 469,395 protein entries from Swiss-Prot [29] database, including all entries available before 2023-02-22. We extracted protein text-based information and protein sequence from Swiss-Prot. Each entry in the dataset is a pair consisting of a protein description and its corresponding protein sequence. The protein description texts consist of three components: protein functions, subcellular localization, and protein families. These characteristics are described using expressive biomedical texts [45]. To identify the correlations between sequence fragments and description sentences, the model preprocesses the two modalities using different tokenizers.

The texts are processed using pre-trained PubMedBERT [31] to obtain description embeddings. The tokenizer of PubMedBERT has a vocabulary size of 30,522, which covers most biomedical terms, enabling effective text embedding with a dimension of 768. For protein sequences, the ESM1b [32] tokenizer is applied for tokenization, as the ESM series models are proficient in capturing the relationships between sequences. Protein sequences are made up of combinations of amino acids, which makes it difficult to identify whether specific groups of amino acids have particular biological functions. To address this challenge, each amino acid is tokenized individually, which maximizes the model's ability to learn the relationship between the entire protein sequence and the description text. This approach enhances the likelihood of identifying protein sequence fragments with specific functions during the protein generation. To ensure uniformity in batch processing, the length of the text is limited to 512 tokens, and the length of the protein sequence is limited to 1024 tokens. The masking pretraining process for the model is outlined in MCM module.

Additionally, ProtDAT does not impose a strict format requirement on the dataset, as long as it is constructed based on specific description content tailored to different tasks. Here, we provide a method for extracting protein texts from Swiss-Prot to construct the dataset. During the training of ProtDAT, the format of the input ProtDAT-Dataset in **Fig 2b** is shown in **Table A1** of **Appendix A**. The protein description texts primarily include protein function, subcellular localization, and protein family information, while the protein sequence is represented by its amino acid tokens. Each sequence has at least one of the aforementioned textual annotations. The specific distribution of all data pairs is shown in **Table A2** of **Appendix A**, which indicates that these three types of description texts have a relatively high proportion in the overall dataset.

**Decoder-Only Masking Pretraining Framework of ProtDAT**

The ProtDAT pretraining framework is based on the decoder-only architecture and trained using the autoregressive generation loss function [46] (denoted as $\mathcal{L}$). It is constructed through 12 decoder layers in **Fig 2c**, where each layer utilizes different attention mechanisms of different types of data. The aim is to embed multimodal information to guide the generation of protein sequences.

(1) **Input embeddings of ProtDAT.** In **Fig 2b**, assume $s = [s_1, s_2 ..., s_n]$ represents the protein sequence tokens obtained after the ESM1b tokenization, and $t = [t_1, t_2 ..., t_m]$ represents the corresponding protein text annotation tokens obtained from PubMedBERT. Since the framework is designed to generate protein sequences with the aid of text, the output of the final layer of the PubMedBERT model $E_{t\_input} = [E_{t_1}, E_{t_2} ..., E_{t_m}]$ is the protein text embedding. Due to the use of a causal mask during sequence training, ProtDAT does not have access to the full information of the protein sequence. Therefore, it is not suitable to directly input the entire sequence into the ESM1b model to obtain embedding vectors. In ProtDAT, after mapping through an embedding layer, the sequence embedding is denoted as $E_{s\_input} = [E_{s_1}, E_{s_2} ..., E_{s_n}]$. Additionally, at the training step of each batch, a cross-modality tensor $c = [c_1, c_2 ..., c_{c\_size}]$ is initialized (where $c\_size$ represents the length of $c$ in ProtDAT).

Upon transformation via the shared embedding layer, this vector is denoted as $\boldsymbol{E_{c\_input}} = [E_{c_1}, E_{c_2} ..., E_{c_{c\_size}}]$. The above three components serve as the protein sequence embedding inputs for the subsequent training process.

(2) **Output embeddings of each decoder layer.** In **Fig 2c**, input embeddings are first passed through layer normalization before being fed into the attention module. Compared to absolute position encoding, the relationship between the sequence and the text is more closely relevant to relative position and neighborhood information. Therefore, during the attention matrix multiplication process, RoPE [30] is applied. After the embeddings pass through the multi-modal cross-attention mechanism (MCM) module, they are then passed through a feedforward network followed by another layer normalization. The output of the $i$-th decoder layer is given by.

$$\boldsymbol{E_s^i, E_c^i, E_t^i} = \left[DL\left(\boldsymbol{E_{s\_input}}, \boldsymbol{E_{c\_input}}, \boldsymbol{E_{t\_input}}\right)\right]_{\times i} \quad (1)$$

where $DL$ represents the decoder layer, $\boldsymbol{E_s^i, E_c^i, E_t^i}$ are also the input of the ($i$+1)-th layer. The final output of the 12-layer decoder consists of the vectors denoted as: $\boldsymbol{E_{s\_output}}, \boldsymbol{E_{t\_output}}, \boldsymbol{E_{c\_output}}$, which respectively represent the embeddings of the protein sequences, the protein annotations, and the modality cross vectors.

(3) **Loss function design.** The loss function is the commonly used next-token prediction loss in autoregressive generative models, which is well-suited for sequential data generation tasks. Since ProtDAT is designed specifically for text-guided protein sequence generation, this loss function ensures that each token prediction aligns with the context provided by preceding tokens, enhancing sequence coherence. Therefore, the loss function is computed only for $\boldsymbol{E_{s\_output}}$.

$$\mathcal{L}(x) = -\sum_{i=1}^{t} \log P(x_i | x_{1:i-1}) \quad (2)$$

$$Loss = \mathcal{L}(\boldsymbol{E_{s\_output}}, s) \quad (3)$$

where $\mathcal{L}$ represents the loss for generating the sequence at time step $t$ based on the sequence from time steps 0 to $t-1$, where it is implemented using the cross-entropy function.

Under this pretraining paradigm, the loss function does not directly involve $\boldsymbol{E_{t\_output}}$ and $\boldsymbol{E_{c\_output}}$. The MCM module of each layer enables the fusion of the two modalities during training, ensuring that the sequence embedding incorporates information from its corresponding protein description text.

To handle variable-length sequences, ProtDAT imposes length limits using padding tokens to standardize the length within each batch. The protein generation model is trained for 200 epochs on 4 A100 GPUs for 5 days. The loss curves for the training and validation sets are shown in **Fig B** of **Appendix B**.

ProtDAT is trained on 469,395 non-redundant protein text-to-sequence description pairs, with a parameter count of 279M. The sequence data represent the arrangement of amino acids within protein molecules, while the textual data consists of information on protein functions, subcellular localization, and protein families. Details of training settings can be found in **Appendix C**.

**ProtDAT Multi-Modal Cross-Attention Mechanism**

To integrate information of diverse modalities, ProtDAT incorporates a multi-modal cross-attention mechanism (MCM) specifically designed for protein sequences and their corresponding descriptive texts, as illustrated in **Fig 2d**. MCM builds effective links for the transformation of multimodal data. Unlike other cross-attention mechanisms [12,16], which directly utilize the query from one modality to interact with the key and value from another modality, MCM implements a different strategy. In protein sequences, each token represents an amino acid, and simply aligning medical text vocabulary with individual amino acid token makes it challenging to learn the relationships between them. Therefore, merely crossing the two modalities is unlikely to yield optimal results, especially in protein generation tasks where error accumulation during the generation process may cause serious consequences. To address this issue, we introduced the **C**ross-**M**odality **I**nteraction **M**odule (CIM) in MCM module to facilitate the nuanced transfer of information from natural language into the protein sequences.

MCM is built upon **M**ulti-**H**ead **A**ttention (MHA) and incorporates the CIM. Additionally, we modified the attention masking mechanism specifically for the CIM, embedding it into the framework to ensure that the interaction between the sequence and textual information. This approach addresses the challenge faced by previous models [15,16], which requires pretraining textual annotations and sequences as separate modules. The MCM module includes three components: **P**rotein **S**equence **M**odule (PSM), **P**rotein **T**ext **M**odule (PTM), and the CIM. During training, the presence of CIM necessitates masking both the descriptive annotations and the protein sequences. For PSM, the token at time step $t$ should only be able to attend to sequences from time steps 0 to $t-1$, so a causal mask should be deployed. For PTM and CIM, since they utilize self-attention and basic cross-attention mechanism, only a padding mask is required.

In **Fig 2d**, $E_s^{i-1}$, $E_t^{i-1}$ and $E_c^{i-1}$ are the inputs to the *i*-th decoder layer, while $E_{s\_input}, E_{t\_input}, E_{c\_input} = E_s^0, E_t^0, E_c^0$. After applying layer normalization and RoPE, linear transformation is used to obtain the Query $Q_t^i$, Key $\mathcal{K}_t^i$, and Value $\mathcal{V}_t^i$ for PTM, and the Query $Q_c^i$ for CIM.

$$E_t^i = \text{softmax}\left(\frac{Q_t^i(\mathcal{K}_t^i)^T}{\sqrt{d_k}}\right)\mathcal{V}_t^i, \quad E_c^i = \text{softmax}\left(\frac{Q_c^i(\mathcal{K}_t^i)^T}{\sqrt{d_k}}\right)\mathcal{V}_t^i \quad (4)$$

where $E_t^i$ represents the output of PTM from the self-attention mechanism at the current layer. $E_c^i$ is the output of CIM from cross-attention mechanism at the *i*-th decoder layer, which contains the compressed and unified text information, preparing for subsequent interaction with the protein sequence modality. Then MCM combines $E_c^i$ to obtain $\mathcal{K}_c^i$ and $\mathcal{V}_c^i$, the updated representations of the sequence and annotation for the next layer are:

$$\mathcal{K}_c^i = Linear_{k\_c}^i(E_c^i), \quad \mathcal{V}_c^i = Linear_{v\_c}^i(E_c^i) \quad (5)$$

where $Linear_{k\_c}^i$ and $Linear_{v\_c}^i$ represent the linear layers that projected from $E_c^i$ to the key and value for CIM.

Subsequently, the MCM module combines the text and sequence information, where $Q_s^i$, $\mathcal{K}_s^i$ and $\mathcal{V}_s^i$ are the query, key, and value for PSM:

$$E_s^i = \text{softmax}\left(\frac{Q_s^i \cdot \left(\text{concat}(\mathcal{K}_c^i, \mathcal{K}_s^i)\right)^T}{\sqrt{d_k}}\right) \cdot \text{concat}(\mathcal{V}_c^i, \mathcal{V}_s^i) \quad (6)$$

where $E_s^i$ serves as the output of PSM within the CCA mechanism. It appropriately integrates the data of different modalities, making it a crucial factor for calculating the loss function and designing new protein sequences. Finally, $E_s^i$, $E_c^i$ and $E_t^i$ are passed through residual connections, layer normalization, and a feedforward layer, formed as the inputs for the (*i*+1)-th decoder layer.

The MCM module is built upon the traditional attention mechanism, centering around CIM to link PSM and PTM. This design successfully incorporates multimodal information, enabling the model to identify associations between biomedical annotations and protein sequences. It also addresses the challenges faced by traditional cross-attention mechanisms, which result in a lack of comprehensive guidance from protein annotations throughout the generation process, thereby limiting the model's ability to accurately capture functional and structural nuances. Notably, to prevent CIM from becoming overly complex and negatively impacting the training process, vector needs to be reinitialized for each batch during the training process.

**Protein Sequence Generation Module of ProtDAT**

To assess the ability to generate protein sequences and design novel proteins, the generation module is specifically constructed for an autoregressive approach: it generates protein sequences one token at one time step, from left to right, with each subsequent token being conditioned on all previously generated tokens. Assume $[x_1, x_2 ... x_{t-1}]$ is the tokenized sequence at time step $t-1$ and text information is $\mathcal{T}$, the calculation process of the probability of generating the protein sequence token at time step $t-1$ is:

$$p(x_t) = \prod_{i=1}^{t} p(x_i | (x_1, x_2 ... x_{i-1}, \mathcal{T})) \tag{7}$$

Due to the relatively small vocabulary size of protein sequences, Top-k parameter is unnecessary for decoding. Instead, a temperature coefficient is applied to control the diversity of the sequence generation process. Additionally, a Top-p decoding strategy is employed to eliminate tokens with very low probabilities, thereby reducing error accumulation. Finally, a repetition penalty parameter is applied to prevent the model from generating a large number of consecutive repetitive tokens during the generation process.

In the case of greedy search [47], the generated sequences tend to have higher repetition and lower similarity to natural sequences. Beam search [48], to some extent, can lead to error accumulation, as the autoregressive generation mechanism relies on previously generated sequences to guide subsequent token selection. Therefore, the model opts for a generation method based on probabilistic sampling [33], where multiple samples are drawn at the start of the procedure to further improve the diversity and orderliness of the generated sequences. In summary, ProtDAT employs a generation strategy [49] that combines Top-p, temperature coefficient, and repetition penalty.

ProtDAT is able to design protein sequences either individually or in batches, the required input prompt can be provided in the following two modes.

**Mode I. Text-Only Input**. ProtDAT only receives the protein descriptive text as input, which can be organized by selecting one or more aspects in protein function, subcellular localization, and protein family information. ProtDAT will generate a completely new protein sequence from scratch with the characteristics described in the given input text.

**Mode II. Text and Sequence Input**. In addition to the input described in Mode I, if there are specific requirements for the generated protein sequence, a prompt fragment can be added for a protein sequence. ProtDAT will then generate the sequence based on the provided protein description text and the input protein sequence fragment. Examples are provided in **Table 4**, while pseudocodes are in **Appendix D**.

**Table 4. Prompt modes in ProtDAT generation process.** Utilizing the ProtDAT framework for protein sequence generation tasks, Mode II is built upon Mode I by providing an additional sequence prompt. As a result, the proteins generated in Mode II are more precise and better aligned with the specific requirements compared to those in Mode I.

| Generating mode | Prompt | Generate Sequence |
| --- | --- | --- |
| Mode I | **Text**=FUNCTION: Is involved in the catabolism of quinate. Allows the utilization of quinate as carbon source via the beta-ketoadipate pathway. SIMILARITY: Belongs to the type-II 3-dehydroquinase family.<br><br>**Sequence**=None | MKACILPINGPNTNLLPTREPVVYGSTTIADVAQATQAASPK… |
| Mode II | **Text**=FUNCTION: Is involved in the catabolism of quinate. Allows the utilization of quinate as carbon source via the beta-ketoadipate pathway. SIMILARITY: Belongs to the type-II 3-dehydroquinase family.<br><br>**Sequence**=MAARILLIN | **MAARILLIN**GPNLNLLGTREPSVYGSTTLADVVAQAKTQAASLN… |

# Acknowledgements

This work was supported by the National Natural Science Foundation of China (No. 62073219).

# Appendix

## A. Details of ProtDAT-Dataset

ProtDAT-Dataset comprises two modalities: protein sequences and protein description texts. **Table A1** presents a few examples, while **Table A2** provides statistical analysis of the entire dataset. Typically, the amino acid vocabulary comprises the standard 25 amino acid names from the IUPAC62 classification.

**Table A1. Examples of ProtDAT-Dataset.** The ProtDAT-Dataset consists of protein descriptive text-to-sequence pairs. On the left side are the amino acid sequences of the proteins, and on the right side are the corresponding protein functions, subcellular localizations, and protein family information in sequential order. In total, there are 469,395 non-redundant data pairs.

| Protein Sequence | Protein Description |
|---|---|
| MAVSDRKLQLLDFEKPLAELEDRIEQIRSLSEQNGVDVTDQIAQLEGRAEQLRQEIFSSLTPMQELQLARHPRRPSTLDYIHAISDEWMELHGDRRGYDDPAIVGGVGRIGGQPVLMLGHQKGRDTKDNVARNFGMPFPSGYRKAMRLMDHADRFGLPIISFIDTPAAWAGLEAEQFGQGEAIALNLREMFRLDVPIICTVIGEGGSGGALGIGVGDRLLMFEHSIYSVAPPEACAAILWRDAQEGPQAAEALKITATDLQELGIIDQILPEPPGGAHVNPIKAANIIKTAILSNLEELWRLSPQERRQQRYHKFRNMGIFSQLP | **FUNCTION:** Component of the acetyl coenzyme A carboxylase complex. First, biotin carboxylase catalyzes the carboxylation of biotin on its carrier protein and then the CO group is transferred by the carboxyltransferase to acetyl-CoA to form malonyl-CoA. <br> **SUBCELLULAR LOCATION**: Cytoplasm. <br> **SIMILARITY**: Belongs to the AccA family. |
| MGNRLIRSYLPNTVMSIEDKQNKYNETIEDSKICNKVYIKQSGKIDKQELTRIKKLGFFYSQKSDHEIERMLFSMPNGTFLLTDDATNENIFIVQKDLENGSLNIAKLEFKGKALYINGKDYYSLENYLKTFEDFYKYPLIYNKNK | **FUNCTION**: Plays a role in virus cell tropism, and may be required for efficient virus replication in macrophages. <br> **SIMILARITY**: Belongs to the asfivirus MGF 100 family. |
| MDLPGNDFDSNDFDAVDLWGADGAEGWTADPIIGVGSAATPDTGPDLDNAHGQAETDTEQEIALFTVTNPPRTVSVSTLMDGRIDHVELSARVAWMSESQLASEILVIADLARQKAQSAQYAFILDRMSQQVDADEHRVALLRKTVGETWGLPSPEEAAAAEAEVFATRYSDDCPAPDDESDPW | **FUNCTION:** Required for ESX-1 function. Required for the maintenance of adequate cellular levels of both EspA and EspC. Facilitates EsxA secretion. <br> **SUBCELLULAR LOCATION:** Secreted. |

**Table A2. Statistical analysis of protein texts.** The distribution of protein description text annotations in the ProtDAT-Dataset is analyzed by counting the frequency of each type of annotation (FUNCTION, SUBCELLULAR LOCATION and SIMILARITY) within the overall data.

| Count | FUNCTION | SUBCELLULAR LOCATION | SIMILARITY |
|---|---|---|---|
| | 393,649 | 308,801 | 430,810 |

In a dataset of 469,395 entries, the proportions of FUNCTION, SUBCELLULAR LOCATION, and SIMILARITY annotations are 83.86%, 65.79%, and 91.78%, respectively. These high proportions indicate a substantial amount of data for each category, suggesting that the training data is well-distributed and less likely to suffer from significant bias issues.

**B. Loss Curves**

The training loss curve is calculated based on $\mathcal{L}$ through the ProtDAT-Dataset training and validation datasets which contain 402,395 and 47,000 protein description text-sequence pairs, respectively. The results are shown in **Fig B**.

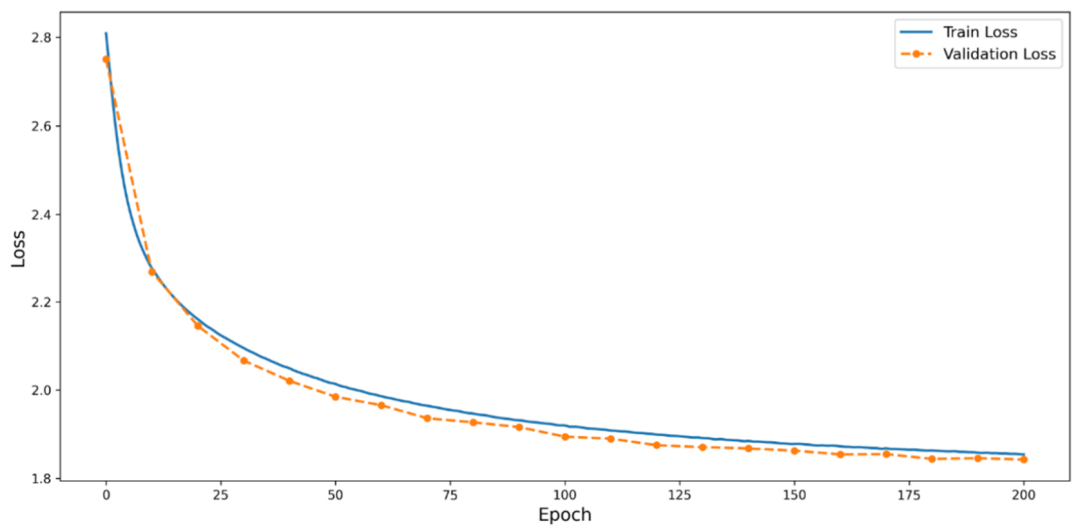

**Figure B**. **Loss curves.** The training and validation dataset of ProtDAT-Dataset loss curves of ProtDAT.

**C. Detailed Training Parameters**

In addition to the training process, Distributed Data Parallel (DDP) mixed-precision training is utilized, employing the AdamW optimizer with a constant learning rate of 1.5e-5. The optimizer parameters are specified as follows. The values of betas are (0.9, 0.999), epsilon is 1e-9, and the weight decay is 0.1. The vector embedding dimension is set to 768, with 12 layers and 12 attention heads. In addition, the batch size of training data on each GPU is 10.

## D. Pseudocodes of Different Generation Modes

Here are the pseudocodes of two generation modes in the generation process, separately named Mode I and Mode II. While $Select$ is the generation strategy of ProtDAT and $s_{prompt}$ denotes the protein sequence prompt fragment.

---
**Mode I:** Only protein text description prompts

**Model:** ProtDAT framework
**Input:** Protein descriptive text (e.g., function, subcellular localization, protein family)
**Output:** Generated protein sequence
    $E_t = PubMedBERT(t)$                                        // Get text embeddings
    $s = [<CLS>]$                                                 // Initialize protein sequence
    while $i < max\_length$ and $s_i\ !=\ <EOS>$:           // Generation stop strategy
        $E_{s\_i} = Model(E_t, E_s, E_c)$
        $s_i = Select(E_{s\_i})$                              // Select amino acid token
        $s.append(s_i)$
    end
    return $s$

---

---
**Mode II:** Protein sequence and text description prompts

**Model:** ProtDAT framework
**Input:** Protein descriptive text (e.g., function, subcellular localization, protein family)
        Protein sequence fragment (e.g., MALS)
**Output:** Generated protein sequence
    $E_t = PubMedBERT(t)$                                        // Get text embeddings
    $s = [<CLS> + s_{prompt}]$                              // Initialize protein sequence
    while $i < max\_length$ and $s_i\ !=\ <EOS>$:           // Generation stop strategy
        $E_{s\_i} = Model(E_t, E_s, E_c)$
        $s_i = Select(E_{s\_i})$                              // Select amino acid token
        $s.append(s_i)$
    end
    return $s$

---

## E. Supplementary of Results with Evaluation Metrics

To further demonstrate ProtDAT's evaluation metrics of **Fig 3e** and **Fig 3f**, results are as follow. In **Figure E(a)**, it reveals that most sequences are concentrated within an

RMSD range between 0 to 3 Å and a pLDDT range of 70% to 100%, further confirming the minimal deviation from the experimental reference structures. Also, **Figure E(b)** shows that most sequences are concentrated within an RMSD range of 0 to 3 Å and a TM-score range between 0.8 to 1.0, indicating a high structural similarity.

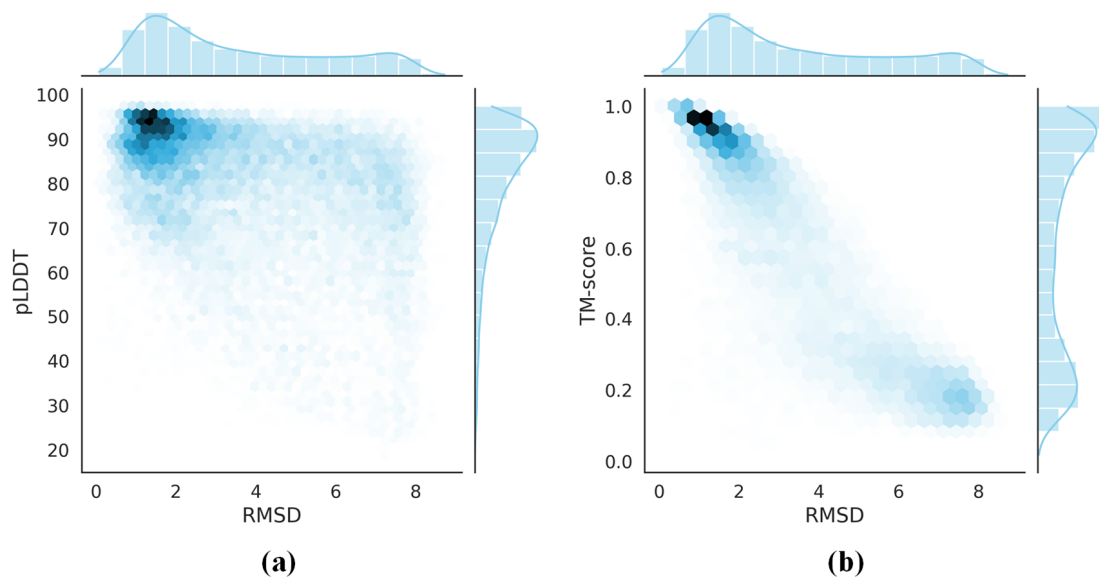

(a)                    (b)

**Figure E**. **Additional distributions of evaluation metrics.** In Figure E, darker colors typically indicate higher data density, meaning more data points in that area. **(a)** The distribution of RMSD and pLDDT. **(b)** The distribution of RMSD and TM-score.